\documentclass[sigconf,screen=true,anonymous=false,bookmarks=false, nonacm]{acmart}
\AtBeginDocument{%
  }

\copyrightyear{2026}
\acmYear{2026}
\setcopyright{cc}
\setcctype{by}
\acmConference[DAC '26]{63rd ACM/IEEE Design Automation Conference}{July 26--29, 2026}{Long Beach, CA, USA}
\acmBooktitle{63rd ACM/IEEE Design Automation Conference (DAC '26), July 26--29, 2026, Long Beach, CA, USA}
\acmDOI{10.1145/3770743.3804207}
\acmISBN{979-8-4007-2254-7/2026/07}

\usepackage{amsmath,amsfonts}
\usepackage{graphicx}
\usepackage[normalem]{ulem}  
\usepackage{xspace}
\usepackage{textcomp}
\usepackage{xcolor}
\usepackage{adjustbox}  %mainly for table shrink
\usepackage{algorithmic}
\usepackage{algorithm}
\usepackage{hyperref}
\usepackage{hyperxmp}
\usepackage{multirow}
\usepackage{makecell}
\usepackage{booktabs}
\usepackage{subcaption}
\usepackage[para]{footmisc} %Footnotes on the same line
\usepackage{tabularx}
\usepackage{colortbl}
\usepackage{arydshln}
\usepackage{siunitx}
\usepackage{placeins}

\usepackage{natbib}

\usepackage{caption}
\definecolor{green1}{HTML}{F1F8E9} % Rank 5 
\definecolor{green2}{HTML}{E2EEDD} % Rank 4
\definecolor{green3}{HTML}{D3E4D1} % Rank 3
\definecolor{green4}{HTML}{C4DAC5} % Rank 2
\definecolor{green5}{HTML}{B5CFB9} % Rank 1 
\definecolor{highlightyellow}{HTML}{FFF9C4}
\newcommand{\cmt}[1]{}

\begin{document}

%%
%% The "title" command has an optional parameter,
%% allowing the author to define a "short title" to be used in page headers.
\title[Zero-Hyperparameters Yield Multi-Corner Analysis Via Learned Priors]{
Breaking the Tuning Barrier: Zero-Hyperparameters Yield Multi-Corner Analysis Via Learned Priors \\
}

%%
%% The "author" command and its associated commands are used to define
%% the authors and their affiliations.
%% Of note is the shared affiliation of the first two authors, and the
%% "authornote" and "authornotemark" commands
%% used to denote shared contribution to the research.
%% TODO: Replace with actual author information

\author{Wei W. Xing}
\affiliation{%
  \institution{School of Mathematical and Physical Science, University of Sheffield}
  \city{Sheffield}
  \country{UK}}
\email{w.xing@sheffield.ac.uk}

\author{Kaiqi Huang}
\affiliation{%
  \institution{SZU–UoS Joint Centre for Innovation and Entrepreneurship, College of Mechatronics and Control Engineering, Shenzhen University
  % \city{Shenzhen}
  \country{China}}
  }

\author{Jiazhan Liu}
\affiliation{%
  \institution{SZU–UoS Joint Centre for Innovation and Entrepreneurship, College of Mechatronics and Control Engineering, Shenzhen University
  % \city{Shenzhen}
  \country{China}}
  }

\author{Hong Qiu}
\affiliation{%
  \institution{SZU–UoS Joint Centre for Innovation and Entrepreneurship, College of Mechatronics and Control Engineering, Shenzhen University
  % \city{Shenzhen}
  \country{China}}
  }

\author{Shan Shen}
\authornote{Corresponding author.}
\affiliation{%
  \institution{Nanjing University of Science and Technology
  % \city{Nanjing}
  \country{China}}
  }
\email{shanshen@njust.edu.cn}

%%
%% The abstract is a short summary of the work to be presented in the
%% article.
\begin{abstract}

Yield Multi-Corner Analysis validates circuits across 25+ Process-Voltage-Temperature corners, resulting in a combinatorial simulation cost of $O(K \times N)$ where $K$ denotes corners and $N$ exceeds $10^4$ samples per corner. Existing methods face a fundamental trade-off: simple models achieve automation but fail on nonlinear circuits, while advanced AI models capture complex behaviors but require hours of hyperparameter tuning per design iteration, forming the Tuning Barrier.
We break this barrier by replacing engineered priors (i.e., model specifications) with learned priors from a foundation model pre-trained on millions of regression tasks. This model performs in-context learning, instantly adapting to each circuit without tuning or retraining. Its attention mechanism automatically transfers knowledge across corners by identifying shared circuit physics between operating conditions. Combined with an automated feature selector (1152D to 48D), our method matches state-of-the-art accuracy (mean MREs as low as 0.11\%) with zero tuning, reducing total validation cost by over $10\times$.

\end{abstract}

%%
%% The code below is generated by the tool at http://dl.acm.org/ccs.cfm.
%% Please copy and paste the code instead of the example below.
%%
% \begin{CCSXML}
% <ccs2012>
%  <concept>
%   <concept_id>10010147.10010178.10010179.10010182</concept_id>
%   <concept_desc>Computing methodologies~Optimization algorithms</concept_desc>
%   <concept_significance>500</concept_significance>
%  </concept>
%  <concept>
%   <concept_id>10010583.10010588.10010559</concept_id>
%   <concept_desc>Hardware~Electronic design automation</concept_desc>
%   <concept_significance>500</concept_significance>
%  </concept>
% </ccs2012>
% \end{CCSXML}

% \ccsdesc[500]{Computing methodologies~Optimization algorithms}
% \ccsdesc[500]{Hardware~Electronic design automation}

%%
%% Keywords. The author(s) should pick words that accurately describe
%% the work being presented. Separate the keywords with commas.
% \keywords{Yield Analysis, Importance Sampling, Multi-Corner Validation}

% \AtBeginDocument{%
%   \providecommand\BibTeX{{%
%     Bib\TeX}}}

%%
%% This command processes the author and affiliation and title
%% information and builds the first part of the formatted document.
\maketitle

\vspace{-0.1in}
\section{Introduction}
\label{sec:introduction}

As integrated circuit technology advances, process variations such as intra-die mismatches, doping fluctuations, and threshold voltage shifts become critical design factors. For modern designs with highly replicated structures like SRAM cell arrays, yield analysis is essential. The ultimate challenge is Yield Multi-Corner Analysis (YMCA), where circuits must be validated across $K$ Process-Voltage-Temperature (PVT) corners, often exceeding 25 configurations. This creates a combinatorial cost barrier: naive Monte Carlo simulation requires $O(K \times N)$ evaluations, where $N > 10^3$ samples per corner are needed for acceptable accuracy. For a 32-transistor SRAM across 25 corners, this translates to over 25,000 SPICE simulations requiring weeks of computation, making iterative design impractical.

The field's pursuit of acceleration has followed two main paths, both reaching fundamental barriers. Importance Sampling (IS) methods like MNIS~\cite{MNIS} achieved remarkable 100 $\times$ speedups through automated norm-minimization, becoming an industry standard. However, simple Gaussian priors create a model capacity barrier: single-point assumptions cannot capture complex, nonlinear failure regions of modern circuits~\cite{EFIAL}. Subsequent clustering approaches~\cite{HSCS,ACS} remained limited by their underlying strong model assumptions.

The second path pursued expressive surrogate-based acceleration~\cite{PEM}. Methods based on Gaussian Processes~\cite{AYEBO}, deep kernels~\cite{ASDK}, and normalizing flows~\cite{nf} successfully broke the model capacity barrier by learning complex, nonlinear failure boundaries. However, this power comes at a steep cost: careful hyperparameter tuning, including kernel selection and network architecture search. The SOTA performance, to some extent, is achieved by careful tuning for particular benchmark problems and is unable to generalize. Thus, these methods are rarely implemented in industrial environments.

\begin{table}[t]
\centering
\footnotesize
\caption{Hyperparameter sensitivity of relative error and \# sim. on 8$\times$2 SRAM with 10 configurations ($\pm20\%$ variation).}
\vspace{-0.15in}
\label{tab:hyper-tuning-summary}
\begin{tabular}{l r| c c c c}
\toprule
 &        & \textbf{OPT}~\cite{nf} & \textbf{FUSIS}~\cite{svmis} & \textbf{ACS}~\cite{ACS} & \textbf{HSCS}~\cite{HSCS} \\
\midrule
\multirow{4}{*}{\rotatebox[origin=c]{90}{MRE (\%)}}
 & Mean        & 45.25  & 79.36  & 69.18  & 74.94  \\
 & Std. Dev.   & 28.96  & 102.43 & 37.75  & 44.60  \\
 & Best (Min)  & 19.36  & 12.35  & 11.92  & 11.11  \\
 & Worst (Max) & 111.24 & 346.32 & 100.00 & 137.78 \\
\midrule
\multirow{4}{*}{\rotatebox[origin=c]{90}{\#Sim.}}
 & Mean        & 461k & 26k & 147k & 195k \\
 & Std. Dev.   & 56k  & 5k  & 65k  & 99k  \\
 & Best (Min)  & 343k & 21k & 42k  & 103k \\
 & Worst (Max) & 500k & 34k & 245k & 437k \\
 \midrule
 & \multicolumn{1}{r|}{\textbf{\# HyperParam.}} & \textbf{10} &\textbf{7} &\textbf{6} &\textbf{3}\\
\bottomrule
\end{tabular}
\vspace{-0.25in}
\end{table}

Table~\ref{tab:hyper-tuning-summary} illustrates the severity of this Tuning Barrier. We evaluate state-of-the-art methods on the 8$\times$2 SRAM benchmark with 10 hyperparameter configurations sampled within $\pm20\%$ of recommended values. Results reveal extreme sensitivity: OPT~\cite{nf} exhibits errors ranging from 19.36\% to 111.24\%, while HSCS~\cite{HSCS} ranges from 11.11\% to 137.78\%. Even the simulation budget varies wildly—ACS~\cite{ACS} consumes between 42k and 245k samples depending on hyperparameter choices. This instability renders these methods unreliable for industrial deployment, where engineers cannot afford hours of expert tuning per design iteration only to face unpredictable performance. This barrier is compounded by multi-corner requirements: while methods like BI-BD~\cite{BI-BD} and BI-BC~\cite{BI-BC} exploit cross-corner correlation, they model only binary Pass/Fail outcomes, discarding continuous performance information that is critical for accurate yield estimation.

We resolve this impasse by replacing engineered priors with learned priors from meta-learning. Traditional methods encode priors through hand-crafted choices: GP kernels specify smoothness assumptions, while IS methods use Gaussian assumptions that sacrifice accuracy. Our framework employs TabPFN~\cite{tabpfn-citation-needed}, a foundation model pre-trained on millions of regression tasks to learn optimal priors. When presented with circuit simulation data, TabPFN performs in-context Bayesian inference through a single forward pass, requiring zero hyperparameter optimization. The learned prior, encoded in pre-trained weights, adapts automatically through attention that identifies and exploits correlations across corners. This cross-corner knowledge transfer is ideal for YMCA: when predicting yield at a sparsely sampled corner, the model borrows strength from well-sampled correlated corners, dramatically improving efficiency.

Our framework integrates this zero-hyperparameter engine into a complete YMCA pipeline. An automated feature selection pipeline discovers sparse, physically interpretable parameter subsets, compressing 1152-dimensional (1152D) circuits to approximately 48 dimensions through single-pass training on initial samples. A global surrogate $\hat{P}(x, c)$ jointly models all corners, with uncertainty-driven active learning focusing simulations on decision boundaries where yield predictions matter most. Our contributions are:

\begin{itemize}
    \item We identify the Tuning Barrier as the fundamental trade-off between model expressiveness and automation that has blocked industrial adoption of modern AI methods for yield analysis.
    
    \item We introduce a learned prior framework using foundation models to achieve expressive modeling with zero per-circuit tuning through in-context learning.
    
    \item We demonstrate automatic cross-corner knowledge transfer, with ablation studies showing over 70\% error reduction on challenging corners through the learned prior.
    
    \item We develop a complete pipeline integrating automated feature selection (1152D to 48D) with zero-hyperparameter inference, achieving state-of-the-art accuracy (mean MREs as low as 0.11\%) and robustness across volatile yield corners, while reducing total validation cost by over $10\times$.
\end{itemize}
    
\vspace{-0.1in}
\section{The Barriers of Yield Analysis}
\label{sec:background}

\subsection{Yield Multi-Corner Estimation}

Consider a circuit characterized by a $D$-dimensional process variation vector $\mathbf{x} = [x_1, \ldots, x_D]^T \in \mathcal{X}$. After standardization, components follow independent Gaussian distributions. The circuit must be validated across $K$ PVT corners $\mathcal{C} = \{c_1, \ldots, c_K\}$, where each corner $c_k \in \mathbb{R}^p$ specifies voltage and temperature conditions.
Performance at process point $\mathbf{x}$ under corner $c_k$ is determined by SPICE simulation $y_k = f(\mathbf{x}, c_k)$. The circuit passes if $f(\mathbf{x}, c_k) > \text{Spec}_k$. Yield at corner $k$ is
\begin{equation}
Y_k = \int_{\mathcal{X}} \mathbb{I}[f(\mathbf{x}, c_k) > \text{Spec}_k] \, p(\mathbf{x}) \, d\mathbf{x},
\end{equation}
where $\mathbb{I}[\cdot]$ denotes the indicator function. Standard Monte Carlo estimation requires $N > 1000$ samples per corner for 3\% relative error. For $K = 25$ corners, this demands over 25,000 SPICE runs with prohibitive $O(K \times N)$ cost.

\vspace{-0.1in}
\subsection{Automation with Model Capacity Barrier}

Importance Sampling methods replace the original distribution $p(\mathbf{x})$ with a proposal $q(\mathbf{x})$ concentrated on failure regions~\cite{HSCS, AIS}. MNIS~\cite{MNIS} achieves massive speedups using a Gaussian proposal constructed from a single minimum-norm failure point. This engineered prior enables full automation. More recent methods like EFIAL~\cite{EFIAL} improve upon MNIS by utilizing all failure samples rather than one, maintaining zero tuning cost.

However, these methods face a fundamental model capacity barrier. MNIS and EFIAL remain constrained by their Gaussian covariance assumptions, which cannot represent multi-modal or highly nonlinear failure boundaries. This creates systematic bias: accuracy plateaus at high error regardless of additional samples. As noted in~\cite{EFIAL}, simple models overlook rich information from failure samples and cannot capture complex failure regions of modern circuits. This is the Model Capacity Barrier.

\vspace{-0.1in}
\subsection{Expressiveness with Tuning Barrier}

Surrogate-based methods overcome capacity limitations by constructing expressive approximations of the SPICE function or failure boundaries. Gaussian Processes~\cite{AYEBO}, deep kernels~\cite{ASDK}, and normalizing flows~\cite{nf} successfully broke the Model Capacity Barrier by capturing arbitrary nonlinearities through flexible model families. Recent work~\cite{svmis} demonstrates that SVM surrogates can construct quasi-optimal IS proposals, unifying surrogate and IS acceleration paths.

However, this expressiveness requires extensive hyperparameter optimization. A 100-dimensional circuit requires optimizing 100+ kernel lengthscales, architecture choices, and training hyperparameters through non-convex optimization, consuming hours per iteration. Critically, this tuning must be repeated for every design change. Table~\ref{tab:hyper-tuning-summary} quantifies this instability: state-of-the-art methods exhibit errors varying from 19\% to 111\% and simulation budgets from 42k to 245k samples under modest $\pm20\%$ hyperparameter perturbations. This per-circuit tuning tax has blocked industrial adoption despite academic success. The field thus faces an impossible trade-off: methods achieve either automation (IS methods) or expressiveness (surrogate methods), but not both—this is the Tuning Barrier.

\vspace{-0.1in}
\section{Breaking the Tuning Barrier via Learned Priors}
\label{sec:framework}

The Tuning Barrier arises from reliance on engineered priors: hand-crafted model specifications that are either too rigid (causing bias) or too flexible (requiring hyperparameter optimization). We replace this foundation with learned priors from meta-learning, eliminating per-circuit tuning while maintaining model expressiveness through a theoretically justified framework. Our approach operates in two stages: sparse feature selection identifies critical parameters, then a zero-hyperparameter inference engine performs efficient multi-corner yield estimation.

\begin{figure}[tb]
    \centering
    \includegraphics[width=1\columnwidth]{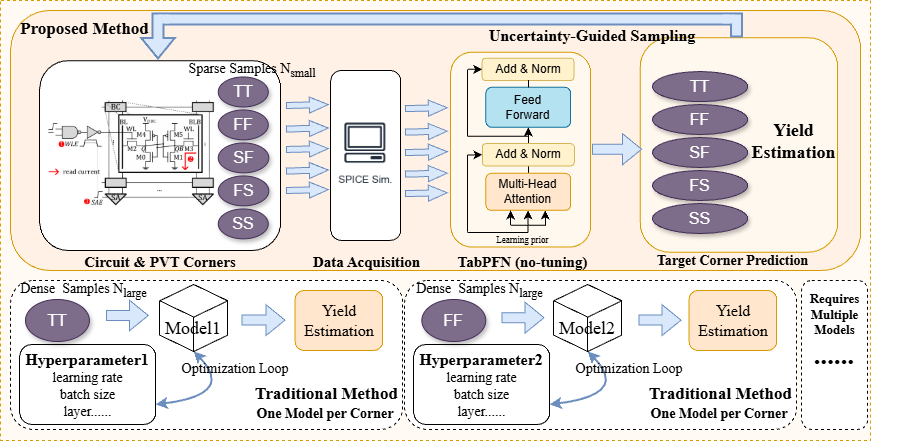}
    \vspace{-0.25in}
    \caption{Automated YMCA framework. Sparse feature selection reduces the problem dimension within TabPFN's capacity. The main loop alternates between in-context learning and uncertainty-driven active sampling until convergence.}
    \label{fig:our_workflow}
    \vspace{-0.2in}
\end{figure}

\vspace{-0.1in}
\subsection{From Engineered to Learned Priors}

The Bayesian posterior predictive distribution (PPD) provides the theoretical foundation for prediction under uncertainty. Given query point $\mathbf{z}^* = [\mathbf{x}^*, c^*]^T$ combining process parameters and corner variables, and training observations $D_{\text{train}} = \{(\mathbf{z}_i, y_i)\}_{i=1}^n$, the PPD integrates over all hypotheses $\phi \in \Phi$ weighted by their posterior probability:
\begin{equation}
p(y^* | \mathbf{z}^*, D_{\text{train}}) = \int_{\Phi} p(y^* | \mathbf{z}^*, \phi) p(\phi | D_{\text{train}}) \, d\phi,
\label{eq:ppd}
\end{equation}
where the posterior $p(\phi | D_{\text{train}}) = p(D_{\text{train}} | \phi) p(\phi) / p(D_{\text{train}})$ follows Bayes' theorem. This integral is intractable for complex hypothesis spaces, necessitating approximation strategies that fundamentally determine the method's expressiveness and automation trade-off.

\textbf{Traditional Approximation: Point Estimates.} Gaussian Processes approximate Eq.~\ref{eq:ppd} by restricting the hypothesis space to a parametric family of kernels $k_{\boldsymbol{\theta}}$ and finding a point estimate $\boldsymbol{\theta}^*$ through marginal likelihood maximization:
\begin{equation}
\boldsymbol{\theta}^* = \arg\max_{\boldsymbol{\theta}} \log p(D_{\text{train}} | \boldsymbol{\theta}) = \arg\max_{\boldsymbol{\theta}} \log \mathcal{N}(\mathbf{y} | \mathbf{0}, K_{\boldsymbol{\theta}} + \sigma^2 I),
\label{eq:gp_mll}
\end{equation}
where $K_{\boldsymbol{\theta}}$ is the Gram matrix with entries $[K_{\boldsymbol{\theta}}]_{ij} = k_{\boldsymbol{\theta}}(\mathbf{z}_i, \mathbf{z}_j)$. For a $D$-dimensional circuit, this requires optimizing $O(D)$ lengthscale parameters plus kernel-specific hyperparameters through non-convex optimization. This constitutes the Tuning Barrier: each new circuit requires hours of optimization, and the point estimate $\boldsymbol{\theta}^*$ ignores uncertainty in hyperparameter selection.

\textbf{Meta-Learning Approximation: Amortized Inference.} We adopt a fundamentally different approximation strategy that amortizes the cost of Bayesian inference across a distribution of functions. Rather than optimize $\boldsymbol{\theta}$ for each circuit, we pre-train a neural network $\mathcal{G}_{\Theta}$ to directly approximate the posterior predictive distribution across a meta-distribution of functions $p_{\text{meta}}(f)$. This Prior-Fitted Network~\cite{tabpfn-citation-needed} is trained once via the meta-learning objective:
\begin{equation}
\small
\Theta^* = \arg\min_{\Theta} \mathbb{E}_{f \sim p_{\text{meta}}(f)} \left[ \mathbb{E}_{D_{\text{train}} \sim f} \left[ \mathbb{E}_{(\mathbf{z}^*, y^*) \sim f} \left[ -\log p_{\Theta}(y^* | \mathbf{z}^*, D_{\text{train}}) \right] \right] \right],
\label{eq:meta_learn}
\end{equation}
where the expectations are taken over: (1) functions $f$ sampled from the meta-prior $p_{\text{meta}}(f)$, (2) training sets $D_{\text{train}}$ sampled from $f$, and (3) test points $(\mathbf{z}^*, y^*)$ sampled from $f$.

\textbf{Theoretical Justification.} The meta-learning objective in Eq.~\ref{eq:meta_learn} can be understood as minimizing the KL divergence between the learned approximation $p_{\Theta}(y^*|\mathbf{z}^*, D_{\text{train}})$ and the true PPD $p(y^*|\mathbf{z}^*, D_{\text{train}})$ averaged over the meta-distribution:
\begin{align}
\small
\Theta^* &= \arg\min_{\Theta} \mathbb{E}_{f, D_{\text{train}}, \mathbf{z}^*}\left[\text{KL}\left(p(y^*|\mathbf{z}^*, D_{\text{train}}) \,\|\, p_{\Theta}(y^*|\mathbf{z}^*, D_{\text{train}})\right)\right] \nonumber
% &= \arg\min_{\Theta} \mathbb{E}_{f, D_{\text{train}}, \mathbf{z}^*}\left[\mathbb{E}_{y^* \sim p(\cdot|\mathbf{z}^*, D_{\text{train}})}\left[-\log p_{\Theta}(y^*|\mathbf{z}^*, D_{\text{train}})\right]\right] + \text{const.}
\label{eq:kl_connection}
\end{align}
This reveals that the meta-learning objective directly minimizes the expected prediction error under the true posterior predictive distribution, providing a principled foundation for the learned prior.
At inference, the trained network performs in-context learning: it accepts new circuit data $D_{\text{circuit}}$ as input and computes the approximate PPD through a single forward pass:
\begin{equation}
(\mu^*, (\sigma^*)^2) = \mathcal{G}_{\Theta^*}(\mathbf{z}^*, D_{\text{circuit}}),
\label{eq:inference}
\end{equation}
where $\mu^*$ and $(\sigma^*)^2$ parametrize the predictive distribution $p_{\Theta^*}(y^*|\mathbf{z}^*, D_{\text{circuit}})$ = $\mathcal{N}(y^*|\mu^*, (\sigma^*)^2)$. Crucially, this requires no gradient descent, no hyperparameter optimization, and no retraining, eliminating the Tuning Barrier while maintaining Bayesian uncertainty quantification.

\textbf{Architecture as Learned Kernel.} We employ TabPFN~\cite{tabpfn-citation-needed}, a Transformer architecture where the self-attention mechanism functions as a learned, nonlinear kernel. For a query point $\mathbf{z}^*$ and training set $D_{\text{circuit}} = \{(\mathbf{z}_i, y_i)\}_{i=1}^n$, the attention mechanism computes:
\begin{equation}
\text{Attention}(\mathbf{z}^*, D_{\text{circuit}}) = \text{softmax}\left(\frac{\mathbf{Q}(\mathbf{z}^*) \mathbf{K}(D_{\text{circuit}})^T}{\sqrt{d_k}}\right) \mathbf{V}(D_{\text{circuit}}),
\label{eq:attention}
\end{equation}
where $\mathbf{Q}$, $\mathbf{K}$, and $\mathbf{V}$ are learned query, key, and value transformations with dimensionality $d_k$. This can be interpreted as a data-dependent kernel $k_{\text{learned}}(\mathbf{z}^*, \mathbf{z}_i; D_{\text{circuit}})$ that adapts to the structure of each new circuit through attention weights:
\begin{equation}
k_{\text{learned}}(\mathbf{z}^*, \mathbf{z}_i; D_{\text{circuit}}) \propto \exp\left(\frac{\mathbf{Q}(\mathbf{z}^*)^T \mathbf{K}(\mathbf{z}_i)}{\sqrt{d_k}}\right),
\end{equation}
Unlike GP kernels that require per-circuit tuning, this kernel is fixed after pre-training yet automatically adapts to new circuits through the attention mechanism's implicit conditioning on $D_{\text{circuit}}$.

\vspace{-0.1in}
\subsection{Cross-Corner Knowledge Transfer}

Rather than modeling each corner $c_k$ independently (which would require $K$ separate models and ignore cross-corner correlation), we formulate a unified global surrogate that learns shared circuit physics across all corners. This design choice enables automatic knowledge transfer between corners, dramatically improving sample efficiency.

\textbf{Joint Input Representation.} We concatenate sparse process parameters $\mathbf{x}_{\mathcal{S}} \in \mathbb{R}^{|\mathcal{S}|}$ with corner encoding $c \in \mathbb{R}^p$ into a unified input vector:
\begin{equation}
\mathbf{z} = \begin{bmatrix} \mathbf{x}_{\mathcal{S}} \\ c \end{bmatrix} \in \mathbb{R}^{|\mathcal{S}|+p},
\label{eq:joint_input}
\end{equation}
where the corner encoding $c$ typically includes normalized voltage and temperature. This enables the global surrogate $\hat{f}(\mathbf{x}_{\mathcal{S}}, c): \mathbb{R}^{|\mathcal{S}|+p} \to \mathbb{R}$ to model performance across all corners simultaneously.

\textbf{Cross-Corner Transfer.} The attention mechanism in Eq.~\ref{eq:attention} automatically identifies and exploits cross-corner correlations. When predicting at a sparsely sampled corner $c_j$ with training data $D_j = \{(\mathbf{z}_i, y_i)\}$ where $\mathbf{z}_i = [\mathbf{x}_i^T, c_j^T]^T$, the attention weights
\begin{equation}
\alpha_{ij} = {\exp\left(\frac{\mathbf{Q}(\mathbf{z}^*)^T \mathbf{K}(\mathbf{z}_i)}{\sqrt{d_k}}\right)}/{\sum_{i'} \exp\left(\frac{\mathbf{Q}(\mathbf{z}^*)^T \mathbf{K}(\mathbf{z}_{i'})}{\sqrt{d_k}}\right)}
\end{equation}
automatically upweight training samples from correlated corners $c_k \approx c_j$ with similar process parameters $\mathbf{x}_k \approx \mathbf{x}^*$. 
The attention mechanism distinguishes corners via the corner encoding $c$, preventing negative transfer between dissimilar corners (e.g., SS and FF).
% This occurs without explicit corner similarity computation: the learned transformations $\mathbf{Q}$ and $\mathbf{K}$ implicitly encode circuit correlations across PVT conditions.

\textbf{Information Sharing Analysis.} The global model enables information flow from well-sampled corners to sparsely sampled ones. Consider two corners $c_1$ (well-sampled, $n_1$ samples) and $c_2$ (sparse, $n_2 \ll n_1$ samples). For a prediction at $(\mathbf{x}^*, c_2)$, the sample size is
\begin{equation}
n_{\text{eff}}(\mathbf{x}^*, c_2) = \sum_{i=1}^{n_1 + n_2} \alpha_{i}^2(\mathbf{x}^*, c_2; D_{\text{all}}) \geq n_2,
\end{equation}
where $D_{\text{all}} = D_1 \cup D_2$ and $\alpha_i$ are normalized attention weights. When corners are correlated (similar circuit physics), samples from $c_1$ receive non-negligible attention weights, effectively increasing $n_{\text{eff}}$ beyond $n_2$ and improving prediction uncertainty. This cross-corner knowledge transfer is particularly valuable for YMCA, where the same circuit topology operates under different PVT conditions that share underlying physical mechanisms.

\subsection{Uncertainty-Guided Active Learning}

The Bayesian nature of TabPFN's learned prior provides calibrated uncertainty estimates $\sigma(\mathbf{z})$ at no additional computational cost—a consequence of approximating the full posterior predictive distribution rather than a point estimate. We exploit this for principled active learning, focusing expensive SPICE simulations on regions where they provide maximum information gain for yield estimation.

\textbf{Information-Theoretic Acquisition Function.} For each corner $c_k$, we define an acquisition function balancing predictive uncertainty with proximity to the specification boundary $\text{Spec}_k$:
\begin{equation}
\alpha_k(\mathbf{x}) = \sigma(\mathbf{x}, c_k) \cdot \phi\left({\hat{f}(\mathbf{x}, c_k) - \text{Spec}_k}/{\sigma(\mathbf{x}, c_k)}\right),
\label{eq:acquisition}
\end{equation}
where $\phi(\cdot)$ is the standard Gaussian density and $\sigma(\mathbf{x}, c_k)$ is the predictive standard deviation from Eq.~\ref{eq:inference}. This uncertainty-weighted boundary sampling has an elegant interpretation: the first term $\sigma(\mathbf{x}, c_k)$ captures epistemic uncertainty (reducible through more data), while the Gaussian density term $\phi(\cdot)$ concentrates sampling near the decision boundary where pass/fail classification is most uncertain. The product achieves maximum value at points that are both highly uncertain and near the specification, where additional samples provide maximum information about yield.

\textbf{Multi-Corner Optimization.} Since yield must be estimated simultaneously across all $K$ corners, we optimize the joint acquisition:
\begin{equation}
\alpha(\mathbf{x}) = \max_{k \in \{1,\ldots,K\}} \alpha_k(\mathbf{x}),
\end{equation}
selecting the corner $k^*$ and process parameters $\mathbf{x}^*$ that maximize information gain across the entire PVT space. For batch parallel simulation (size $B$), we employ a greedy diversification strategy: after selecting $\mathbf{x}_1^*$, subsequent selections maximize $\alpha(\mathbf{x}) - \beta \sum_{b=1}^{j-1} \exp(-\|\mathbf{x} - \mathbf{x}_b^*\|^2 / 2\gamma^2)$ where the penalty term ($\beta, \gamma > 0$) encourages spatial diversity in the selected batch.

\vspace{-0.1in}
\subsection{Scalability via Sparse Feature Selection}

TabPFN's learned prior is most effective for moderate-dimensional spaces ($D \le 100$). Industrial circuits like 32$\times$2 SRAMs with $D=1152$ variation parameters exceed this range. However, circuit physics exhibits inherent sparsity: performance typically depends on a small subset of critical transistors rather than all 1152 parameters. We exploit this through an automated, zero-tuning feature selection pipeline that identifies critical parameters without introducing the hyperparameter optimization burden we seek to avoid.

\textbf{Greedy Forward Selection Framework.} Our approach employs a validation-driven greedy search guided by initial importance ranking. Given initial samples $D_0$, we first train a Gradient Boosting Decision Tree (GBDT) model $M_{\text{gbdt}}$ using default LightGBM configuration—no hyperparameter tuning required. GBDT models naturally produce feature importance scores $\{I_j\}_{j=1}^{D+p}$ as a byproduct of training, quantifying each feature's aggregate contribution across all tree splits. We rank features by importance: $\mathcal{F}_{\text{ranked}} = (f_{(1)}, f_{(2)}, \ldots, f_{(D+p)})$ where $I_{(1)} \ge I_{(2)} \ge \ldots \ge I_{(D+p)}$.

With this ranking established, we split $D_0$ into training ($D_{\text{train}}$, 80\%) and validation ($D_{\text{val}}$, 20\%) subsets. For each candidate subset size $k \in \{1, 2, \ldots, \lfloor(D+p)/B\rfloor\}$ where $B$ is a fixed batch size, we construct $\mathcal{S}_k$ containing the top $k \times B$ features from $\mathcal{F}_{\text{ranked}}$. A fresh GBDT model $M_k$ is trained on $D_{\text{train}}$ using only features in $\mathcal{S}_k$, and evaluated on $D_{\text{val}}$ via $R^2$ score:
\begin{equation}
R^2(\mathcal{S}_k) = 1 - \frac{\sum_{(\mathbf{z}, y) \in D_{\text{val}}} (y - M_k(\mathbf{z}))^2}{\sum_{(\mathbf{z}, y) \in D_{\text{val}}} (y - \bar{y}_{\text{val}})^2},
\label{eq:r2_score}
\end{equation}
where $\bar{y}_{\text{val}}$ is the mean target value in $D_{\text{val}}$. The optimal subset is:
\begin{equation}
\mathcal{S}^* = \arg\max_{k} R^2(\mathcal{S}_k).
\label{eq:feat_select_greedy}
\end{equation}

\textbf{Zero-Tuning Guarantee.} Critically, this entire pipeline requires no hyperparameter optimization. The GBDT models use fixed default configurations optimized for general small-sample scenarios, applied universally across all circuits and corners in our experiments without case-specific adjustments. The batch size $B=10$ is a non-sensitive granularity parameter, not a tunable hyperparameter. Validation performance (Eq.~\ref{eq:r2_score}) serves as the sole selection criterion (Eq.~\ref{eq:feat_select_greedy}), eliminating manual hyperparameter choices. This one-time, sub-minute preprocessing typically compresses a 1152-dimensional SRAM array to approximately 48 dimensions ($|\mathcal{S}^*| \approx 48$), bringing them within TabPFN's effective range while preserving physical interpretability.

This feature selection addresses a limitation of the current TabPFN implementation rather than a fundamental algorithmic constraint. The 500-feature limit stems from TabPFN's training on datasets with up to 500 features---a practical choice given computational resources. If TabPFN were trained on higher-dimensional synthetic datasets (which our SCM/BNN prior can readily generate), it would naturally handle 1152-dimensional circuits directly. Our feature selection thus serves as an efficient preprocessing step for the current model, eliminating the need to retrain TabPFN while maintaining the zero-hyperparameter philosophy throughout the entire pipeline.

Algorithm~\ref{alg:main} summarizes the complete procedure. Computational cost per iteration is dominated by SPICE simulation (minutes to hours); TabPFN inference via Eq.~\ref{eq:inference} and acquisition optimization complete in seconds due to the amortized inference paradigm. The algorithm converges when yield estimates stabilize: $\max_k |\hat{Y}_k^{(t)} - \hat{Y}_k^{(t-1)}| < \epsilon$ for convergence threshold $\epsilon$.

\begin{algorithm}[]
    \caption{Zero-Hyperparameter YMCA Framework}
    \label{alg:main}
    \begin{algorithmic}[1]
    \REQUIRE Corners $\mathcal{C}$, specifications $\{\text{Spec}_k\}$, budgets $N_{\text{total}}, N_{\text{init}}, B$
    \ENSURE Yield estimates $\{\hat{Y}_k\}_{k=1}^K$
    
    \STATE \textbf{Initialize:} $D_0 \leftarrow \text{LHS-Sample-and-Simulate}(N_{\text{init}} \times K)$
    \STATE $\mathcal{S} \leftarrow \{1, \ldots, D+p\}$ \COMMENT{Full feature set}
    
    \IF{$D + p > 500$}
        \STATE $\mathcal{S}^* \leftarrow \text{SelectFeatures}_{\text{Greedy}}(D_0, \text{Eq.~\ref{eq:feat_select_greedy}})$
        \STATE $D_0 \leftarrow \text{Project}(D_0, \mathcal{S}^*)$
        \STATE $\mathcal{S} \leftarrow \mathcal{S}^*$ \COMMENT{Update to the optimal subset}
    \ENDIF
    
    \STATE $D_t \leftarrow D_0$
    \WHILE{$|D_t| < N_{\text{total}}$ and not converged}
        \STATE \textbf{In-Context Learning:} $(\hat{f}_t, \sigma_t) \leftarrow \text{TabPFN}(D_t)$ 
        
        \STATE \textbf{Yield Estimation:} $\hat{Y}_k^{(t)} \leftarrow \text{EstimateYield}(\hat{f}_t, \mathcal{S}, M=10^6)$ 
        
        \STATE \textbf{Active Learning:} $\{\mathbf{x}_b^*, k_b^*\}_{b=1}^B \leftarrow \text{Select}(\hat{f}_t, \sigma_t, \mathcal{S}, \text{Eq.~\ref{eq:acquisition}})$
        
        \STATE \textbf{Simulate \& Update:} $\{y_b^*\}_{b=1}^B \leftarrow \text{SPICE}(\{(\mathbf{x}_b^*, c_{k_b^*})\}_{b=1}^B)$
        \STATE $D_{t+1} \leftarrow D_t \cup \{(\text{Project}(\mathbf{x}_b^*, \mathcal{S}), c_{k_b^*}, y_b^*)\}_{b=1}^B$
    \ENDWHILE
    \RETURN $\{\hat{Y}_k^{(t)}\}_{k=1}^K$
    \end{algorithmic}
\end{algorithm}

\vspace{-0.2in}
\section{Experimental Validation}
\label{sec:experiments}

\noindent With the framework established, we now validate it on industrial SRAM benchmarks. We first outline the benchmark circuits and PVT conditions, then evaluate our method against Monte Carlo and state-of-the-art baselines under consistent metrics, with particular emphasis on total validation cost.

\begin{figure}[]
    \centering
    \vspace{-0.1in}
    \includegraphics[width=0.65\columnwidth]{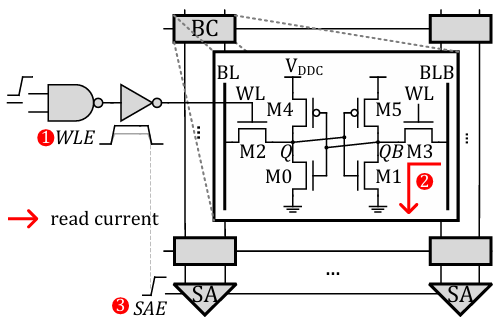} 
    \vspace{-0.15in}
    \caption{Read current path of 6T SRAM cell from OpenYield benchmark suite~\cite{shen2025openyield}.}
    \label{fig:sram_cell}
    \vspace{-0.15in}
\end{figure}

% \subsection{Experimental Setup}

\noindent\textbf{Benchmark Circuits.}
We use OpenYield’s industrial-quality SRAM macros in FreePDK45 (45 nm), which include critical peripherals—sense amplifiers (SA), column mux, wordline drivers, and decoders—and capture second-order effects (parasitics, leakage, and PVT variations)~\cite{shen2025openyield}. As illustrated in Figure~\ref{fig:sram_cell}, a read operation precharges both bitlines and then asserts the wordline to connect the storage node through the access transistor; a read failure occurs when variations prevent sufficient bitline differential within the delay specification or induce a read disturb. We evaluate five PVT corners (TT, SS, FF, FS, SF) across four array sizes: 4$\times$2, 8$\times$2, 16$\times$2, and 32$\times$2. These realistic testcases comprised of full peripheral circuits on the read path highlight the proposed model’s robustness across corner conditions and SRAM sizes. Tables~\ref{tab:pvt_combo} and~\ref{tab:sram_configs} summarize the operating conditions and configurations.

\begin{table}[]
\centering
\caption{PVT Combinations used in our experiments. The process corners in the context represent a PVT combination.}
\vspace{-0.15in}
\label{tab:pvt_combo}
\begin{tabular}{llllll}
\toprule
\textbf{Process Corner}  & \textbf{TT} & \textbf{FF} & \textbf{SF} & \textbf{FS} & \textbf{SS} \\
\midrule
Voltage (V)     & 1.0 & 1.1 & 1.0 & 1.0 & 0.9 \\
Temperature (C) & 25  & 0 & 25 & 25 & 125 \\
\#Simulations   & 10k & 10k & 10k & 10k & 10k \\
\bottomrule
\end{tabular}
\vspace{-0.15in}
\end{table}

\begin{table}[]
\centering
\caption{SRAM configurations used in our experiments. The number of rows spans 4--32, while the number of columns remains 2 for faster simulations.}
\vspace{-0.15in}
\label{tab:sram_configs}
\begin{tabular}{lllll}
\toprule
\textbf{Testcase}        & \textbf{1} & \textbf{2} & \textbf{3}  & \textbf{4}  \\
\midrule
\#Row           & 4 & 8 & 16 & 32 \\
\#Column        & 2 & 2 & 2  & 2  \\
Delay Spec (ns) & 0.0626  & 0.0745 & 0.113 & 0.209 \\
\#Process Para. & 144 & 288 & 576 & 1152 \\
\bottomrule
\end{tabular}
\vspace{-0.15in}
\end{table}

% \subsubsection{Baseline Methods}
\noindent\textbf{Baseline Methods.} 
For YMCA, we compare against Monte Carlo (MC) as the unbiased reference; BI-BD and BI-BC~\cite{BI-BD, BI-BC}, which exploit cross-corner correlation; and OPT~\cite{nf}, a complex-model-driven yield analysis with corner transfer capacity. For single-corner yield analysis, we include Importance Sampling (IS) baselines: the industry-standard MNIS~\cite{MNIS}, and more recent methods HSCS~\cite{HSCS} and ACS~\cite{ACS}. Prior evaluations typically use idealized SRAM bitcells without peripherals or second-order effects; in contrast, our comparisons are run on realistic macro-level OpenYield testcases, ensuring results that reflect practical conditions.

% \subsubsection{Evaluation Metrics}
\noindent\textbf{Evaluation Metrics.} To ensure consistent comparisons, we report accuracy and efficiency using relative error $\text{error} = |\hat{Y} - Y|/Y \times 100\%$ against the golden MC reference, Mean Relative Error (MRE) $MRE = \frac{1}{K}\sum_{k=1}^{K}|\text{error}_{k}|$ across corners, and speedup $S = N_{\text{MC}}/N_{\text{method}}$. Critically, we evaluate the total validation cost by combining simulation time. When convenient, we also use the failure rate $P_{fail}=1-Y$ as an equivalent yield metric.
\vspace{-0.1in}

\subsection{TabPFN versus Traditional Surrogates}

To assess our learned-prior engine's competitiveness, we compare TabPFN against traditional surrogates, including Gaussian Process (GP), Deep-GP, MLP, and SVM, on the 8 $\times$ 2 SRAM benchmark. 
To ensure the comparison reflects fundamental model capacity rather than insufficient effort, all baseline methods were tuned using 5 random search to find the best results, whereas TabPFN was applied with fixed defaults and zero tuning.

\begin{figure}[]
\centering
\includegraphics[width=0.95\columnwidth]{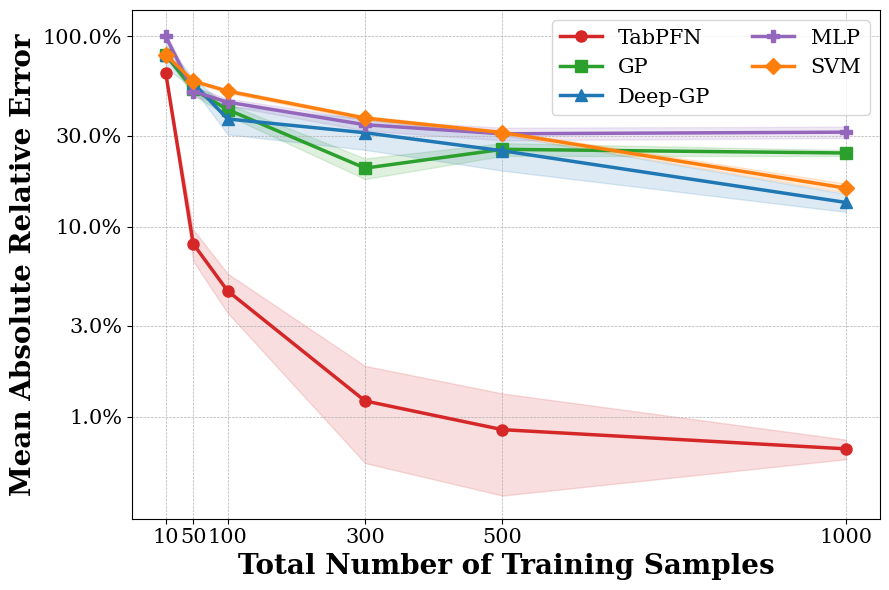} 
\vspace{-0.15in}
\caption{Surrogate model comparison on 8$\times$2 SRAM. TabPFN (red) exhibits superior data efficiency in the small-sample regime while requiring zero hyperparameter tuning.}
\label{fig:surrogate_benchmark}
\vspace{-0.15in}
\end{figure}

Figure~\ref{fig:surrogate_benchmark} shows TabPFN's superior data efficiency in the small-sample regime ($<1000$). Baselines do show early status due to their unstable performance.
Around 100 samples, TabPFN achieves low error ($\sim$5\% MAE), while all tuned baselines lag significantly, with MAE ranging from $\sim$30\% (GP) to over $\sim$45\% (MLP). As the budget increases, all methods improve, but TabPFN maintains a clear margin while eliminating tuning overhead. This confirms that learned priors deliver strong modeling efficiency without per-circuit training, making them well-suited as the core of our framework.

\begin{table*}[t!]
    \centering
    \caption{Detailed yield predictions for all methods across PVT corners. Ground truth results $Y$ (\%) from 50,000-sample MC, predicted yield $\hat{Y}$ (\%), and MRE. The lowest MRE per corner is bolded.}
    \label{tab:multi_corner_results_detailed}
    \small
    \setlength{\tabcolsep}{5.pt}
    \renewcommand{\arraystretch}{1}
    \vspace{-0.15in}
    \begin{tabular}{c l rrr  rrr  rrr  rrr  rrr}
    \toprule
    \multirow{2}{*}{\textbf{Circuit}} & \multirow{2}{*}{\textbf{Method}} & 
    \multicolumn{3}{c}{\textbf{TT}} & \multicolumn{3}{c}{\textbf{FF}} & \multicolumn{3}{c}{\textbf{SF}} & \multicolumn{3}{c}{\textbf{FS}} & \multicolumn{3}{c}{\textbf{SS}} \\
    \cmidrule(lr){3-5} \cmidrule(lr){6-8} \cmidrule(lr){9-11} \cmidrule(lr){12-14} \cmidrule(lr){15-17}
     &   & $Y$ & $\hat{Y}$ &  MRE(\%) & $Y$ & $\hat{Y}$ & MRE(\%) & $Y$ & $\hat{Y}$ & MRE(\%) & $Y$ & $\hat{Y}$ & MRE(\%) & $Y$ & $\hat{Y}$ & MRE(\%) \\
    \midrule
    \multirow{4}{*}{\rotatebox{90}{4$\times$2}}
    & BI-BD & \cellcolor{green5}100 & \cellcolor{green5}100 & \cellcolor{green5}\textbf{0} & \cellcolor{green3}95.0 & \cellcolor{green3}95.4 & \cellcolor{green3}\textbf{0.42} & \cellcolor{green2}90.9 & \cellcolor{green2}91.2 & \cellcolor{green2}0.33 & \cellcolor{green5}100 & \cellcolor{green5}100 & \cellcolor{green5}\textbf{0} & \cellcolor{green1}0 & \cellcolor{green1}0 & \cellcolor{green1}\textbf{0} \\
    & BI-BC & \cellcolor{green5}100 & \cellcolor{green5}100 & \cellcolor{green5}\textbf{0} & \cellcolor{green3}95.0 & \cellcolor{green3}96.4 & \cellcolor{green3}1.47 & \cellcolor{green2}90.9 & \cellcolor{green2}91.6 & \cellcolor{green2}0.77 & \cellcolor{green5}100 & \cellcolor{green5}100 & \cellcolor{green5}\textbf{0} & \cellcolor{green1}0 & \cellcolor{green1}0 & \cellcolor{green1}\textbf{0} \\
    & OPT & \cellcolor{green5}100 & \cellcolor{green5}100 & \cellcolor{green5}\textbf{0} & \cellcolor{green3}95.0 & \cellcolor{green3}95.9 & \cellcolor{green3}0.94 & \cellcolor{green2}90.9 & \cellcolor{green2}92.2 & \cellcolor{green2}1.43 & \cellcolor{green5}100 & \cellcolor{green5}100 & \cellcolor{green5}\textbf{0} & \cellcolor{green1}0 & \cellcolor{green1}0 & \cellcolor{green1}\textbf{0} \\
    & Proposed & \cellcolor{green5}100 & \cellcolor{green5}100 & \cellcolor{green5}\textbf{0} & \cellcolor{green3}95.0 & \cellcolor{green3}95.4 & \cellcolor{green3}\textbf{0.42} & \cellcolor{green2}90.9 & \cellcolor{green2}90.8 & \cellcolor{green2}\textbf{0.11} & \cellcolor{green5}100 & \cellcolor{green5}100 & \cellcolor{green5}\textbf{0} & \cellcolor{green1}0 & \cellcolor{green1}0 & \cellcolor{green1}\textbf{0} \\
    \midrule
    \multirow{4}{*}{\rotatebox{90}{8$\times$2}}
    & BI-BD & \cellcolor{green3}87.1 & \cellcolor{green3}87.4 & \cellcolor{green3}0.34 & \cellcolor{green4}99.0 & \cellcolor{green4}99.2 & \cellcolor{green4}0.20 & \cellcolor{green2}11.3 & \cellcolor{green2}11.2 & \cellcolor{green2}0.89 & \cellcolor{green5}100 & \cellcolor{green5}100 & \cellcolor{green5}\textbf{0} & \cellcolor{green1}0 & \cellcolor{green1}0 & \cellcolor{green1}\textbf{0} \\
    & BI-BC & \cellcolor{green3}87.1 & \cellcolor{green3}88.8 & \cellcolor{green3}1.95 & \cellcolor{green4}99.0 & \cellcolor{green4}98.4 & \cellcolor{green4}0.61 & \cellcolor{green2}11.3 & \cellcolor{green2}12.4 & \cellcolor{green2}9.73 & \cellcolor{green5}100 & \cellcolor{green5}100 & \cellcolor{green5}\textbf{0} & \cellcolor{green1}0 & \cellcolor{green1}0 & \cellcolor{green1}\textbf{0} \\
    & OPT & \cellcolor{green3}87.1 & \cellcolor{green3}88.5 & \cellcolor{green3}1.61 & \cellcolor{green4}99.0 & \cellcolor{green4}99.4 & \cellcolor{green4}0.40 & \cellcolor{green2}11.3 & \cellcolor{green2}29.9 & \cellcolor{green2}100+ & \cellcolor{green5}100 & \cellcolor{green5}100 & \cellcolor{green5}\textbf{0} & \cellcolor{green1}0 & \cellcolor{green1}0 & \cellcolor{green1}\textbf{0} \\
    & Proposed & \cellcolor{green3}87.1 & \cellcolor{green3}87.3 & \cellcolor{green3}\textbf{0.23} & \cellcolor{green4}99.0 & \cellcolor{green4}99.0 & \cellcolor{green4}\textbf{0.01} & \cellcolor{green2}11.3 & \cellcolor{green2}11.4 & \cellcolor{green2}\textbf{0.88} & \cellcolor{green5}100 & \cellcolor{green5}100 & \cellcolor{green5}\textbf{0} & \cellcolor{green1}0 & \cellcolor{green1}0 & \cellcolor{green1}\textbf{0} \\
    \midrule
    \multirow{4}{*}{\rotatebox{90}{16$\times$2}}
    & BI-BD & \cellcolor{green2}31.5 & \cellcolor{green2}36.2 & \cellcolor{green2}14.9 & \cellcolor{green5}100 & \cellcolor{green5}100 & \cellcolor{green5}\textbf{0} & \cellcolor{green1}0.14 & \cellcolor{green1}0 & \cellcolor{green1}100+ & \cellcolor{green4}97.8 & \cellcolor{green4}98.6 & \cellcolor{green4}0.82 & \cellcolor{green3}96.1 & \cellcolor{green3}94.9 & \cellcolor{green3}1.25 \\
    & BI-BC & \cellcolor{green2}31.5 & \cellcolor{green2}32.0 & \cellcolor{green2}1.59 & \cellcolor{green5}100 & \cellcolor{green5}100 & \cellcolor{green5}\textbf{0} & \cellcolor{green1}0.14 & \cellcolor{green1}0.50 & \cellcolor{green1}100+ & \cellcolor{green4}97.8 & \cellcolor{green4}98.3 & \cellcolor{green4}0.51 & \cellcolor{green3}96.1 & \cellcolor{green3}96.8 & \cellcolor{green3}0.73 \\
    & OPT & \cellcolor{green2}31.5 & \cellcolor{green2}45.6 & \cellcolor{green2}44.8 & \cellcolor{green5}100 & \cellcolor{green5}100 & \cellcolor{green5}\textbf{0} & \cellcolor{green1}0.14 & \cellcolor{green1}22.1 & \cellcolor{green1}100+ & \cellcolor{green4}97.8 & \cellcolor{green4}97.9 & \cellcolor{green4}3.07 & \cellcolor{green3}96.1 & \cellcolor{green3}96.9 & \cellcolor{green3}3.85 \\
    & Proposed & \cellcolor{green2}31.5 & \cellcolor{green2}31.7 & \cellcolor{green2}\textbf{0.63} & \cellcolor{green5}100 & \cellcolor{green5}100 & \cellcolor{green5}\textbf{0} & \cellcolor{green1}0.14 & \cellcolor{green1}0.20 & \cellcolor{green1}\textbf{42.8} & \cellcolor{green4}97.8 & \cellcolor{green4}97.5 & \cellcolor{green4}\textbf{0.31} & \cellcolor{green3}96.1 & \cellcolor{green3}96.6 & \cellcolor{green3}\textbf{0.52} \\
    \midrule
    \multirow{4}{*}{\rotatebox{90}{32$\times$2}}
    & BI-BD & \cellcolor{green3}97.6 & \cellcolor{green3}97.9 & \cellcolor{green3}0.31 & \cellcolor{green5}100 & \cellcolor{green5}100 & \cellcolor{green5}\textbf{0} & \cellcolor{green1}34.9 & \cellcolor{green1}48.0 & \cellcolor{green1}37.5 & \cellcolor{green5}100 & \cellcolor{green5}100 & \cellcolor{green5}\textbf{0} & \cellcolor{green2}85.0 & \cellcolor{green2}86.9 & \cellcolor{green2}2.24 \\
    & BI-BC & \cellcolor{green3}97.6 & \cellcolor{green3}98.0 & \cellcolor{green3}0.41 & \cellcolor{green5}100 & \cellcolor{green5}100 & \cellcolor{green5}\textbf{0} & \cellcolor{green1}34.9 & \cellcolor{green1}37.0 & \cellcolor{green1}6.02 & \cellcolor{green5}100 & \cellcolor{green5}100 & \cellcolor{green5}\textbf{0} & \cellcolor{green2}85.0 & \cellcolor{green2}86.5 & \cellcolor{green2}\textbf{1.76} \\
    & OPT & \cellcolor{green3}97.6 & \cellcolor{green3}97.4 & \cellcolor{green3}3.07 & \cellcolor{green5}100 & \cellcolor{green5}100 & \cellcolor{green5}\textbf{0} & \cellcolor{green1}34.9 & \cellcolor{green1}54.1 & \cellcolor{green1}55.0 & \cellcolor{green5}100 & \cellcolor{green5}100 & \cellcolor{green5}\textbf{0} & \cellcolor{green2}85.0 & \cellcolor{green2}87.6 & \cellcolor{green2}3.05 \\
    & Proposed & \cellcolor{green3}97.6 & \cellcolor{green3}97.7 & \cellcolor{green3}\textbf{0.10} & \cellcolor{green5}100 & \cellcolor{green5}100 & \cellcolor{green5}\textbf{0} & \cellcolor{green1}34.9 & \cellcolor{green1}36.0 & \cellcolor{green1}\textbf{3.15} & \cellcolor{green5}100 & \cellcolor{green5}100 & \cellcolor{green5}\textbf{0} & \cellcolor{green2}85.0 & \cellcolor{green2}86.9 & \cellcolor{green2}2.24 \\
    \bottomrule
    \end{tabular}
    \vspace{-0.15in}
\end{table*}

\begin{figure}[]
\centering
\includegraphics[width=1\columnwidth]{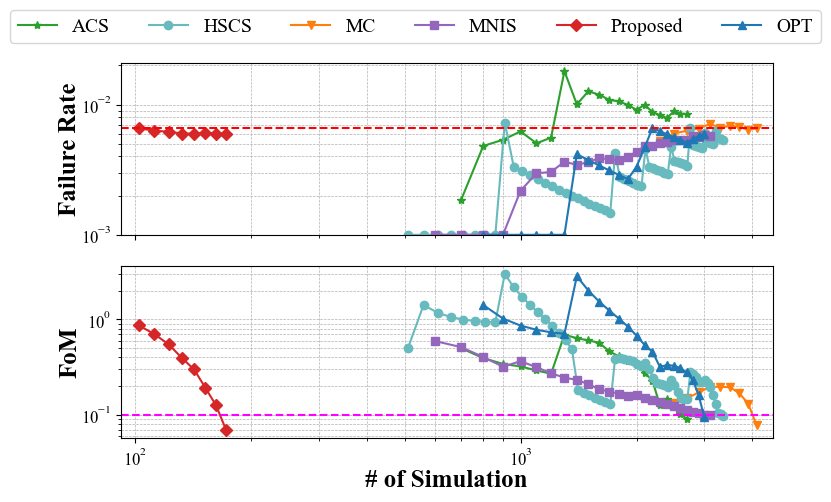} 
\vspace{-0.35in}
\caption{Convergence on single challenging corner (8$\times$2 SRAM, FF corner) of SOTA yield analysis methods.}
\label{fig:single_corner}
\vspace{-0.15in}
\end{figure}

\begin{table}[htb]
\centering
\caption{Single-corner performance on 8$\times$2 SRAM FF corner against 4,100-sample MC baseline.}
\label{tab:model_comparison_k}
\vspace{-0.1in}
\sisetup{
    detect-weight,
    mode = text
}
\begin{tabular}{@{} l S[table-format=1.2e-1, scientific-notation=true] 
                  S[table-format=-2.2] 
                  S[table-format=4.0]
                  S[table-format=2.1, table-space-text-post=x] 
                  @{}}
\toprule
\textbf{Method} & $\boldsymbol{P_{fail}}$ & {\textbf{MRE (\%)}} & {\textbf{\#Sim.}} & {\textbf{Speedup}} \\
\midrule
MC      & 6.59e-3 & {---}    & 4100 & {---} \\
OPT     & 6.00e-03 & \textbf{8.88}   & 3000 & 1.4x \\
MNIS    & 5.75e-03 & 12.63 & 3100  & 1.3x \\
ACS     & 8.46e-3 & 28.47   & 2700 & 1.5x \\
HSCS    & 5.33e-3 & 19.01  & 3360 & 1.2x \\
\midrule
\textbf{Proposed} & \bfseries 6.00e-3 & \bfseries 8.88 & \bfseries 170 & \bfseries 24.1x \\
\bottomrule
\end{tabular}
\vspace{-0.1in}
\end{table}
\vspace{-0.1in}

\subsection{Single-Corner Yield Analysis}

We assess yield analysis in a challenging single-corner setting: the 8$\times$2 SRAM at the FF corner.
This checks whether the proposed method delivers the top performance in a challenging single-corner yield analysis problem.

Figure~\ref{fig:single_corner} illustrates the yield estimation convergence.
IS methods (HSCS, ACS) exhibit pronounced cold-start phases: errors remain high while hundreds of simulations are spent exploring to locate failure regions.
In contrast, our approach avoids this overhead. With a learned prior, TabPFN produces reasonable approximations from about 100 samples, converges below 200 samples. This superior performance allows active learning to refine immediately and yield monotonic error reduction from the start.

Table~\ref{tab:model_comparison_k} reports final performance consistent with the figure. Our zero-tuning framework achieves {8.88\% MRE with 170 simulations}, delivering a {24.1x} reduction relative to the 4,100-sample MC baseline. HSCS requires 3,360 samples for 19.01\% MRE, ACS 2,700 for 28.47\%, MNIS 3,100 for 12.63\%. Notably, our method matches OPT’s accuracy (8.88\% MRE) while using over {17x} fewer simulations.

Importantly, many prior SOTA methods are developed and evaluated on idealized SRAM bitcells or toy circuits. Under non-ideal macro-level conditions, including input skew, large bitline loads, and noise from peripherals and neighbor cells, their gains diminish: in our realistic setup they deliver only {1.2–1.5x} speedups with {12.63–28.47\%} MRE (Table~\ref{tab:model_comparison_k}). In contrast, our learned-prior approach remains accurate and data-efficient in the presence of second-order effects, confirming robustness and practical advantage in yield analysis.
\vspace{-0.1in}

\subsection{Cross-Corner Knowledge Transfer}

\begin{table}[tb]
    \centering
    \caption{Ablation study demonstrating cross-corner knowledge transfer on 16$\times$2 SRAM. Each target corner uses 50 samples from itself, plus 50 samples from an unseen corner.}
    \label{tab:global_modeling_ablation}
    \sisetup{round-mode=places, round-precision=2, table-auto-round=true}
    \vspace{-0.15in}
    \resizebox{\columnwidth}{!}{
    \begin{tabular}{c c c c c c c}
    \toprule
     &  \multicolumn{5}{c}{\textbf{Additional Corners}} & \\
    \cmidrule(lr){3-6}
    \textbf{Corner} & \makecell{\textbf{Target}\\\textbf{Only}} & \makecell{\textbf{+1}} & \makecell{\textbf{+2}} & \makecell{\textbf{+3}} & \makecell{\textbf{+4}} & \makecell{\textbf{MRE}\\ \textbf{Reduction}} \\
    \midrule
    TT & 21.79 & 13.85 & 10.86 & 9.05 & \textbf{6.04} & $-72\%$ \\
    FF & 0.00 & 0.00 & 0.00 & 0.00 & \textbf{0.00} & --- \\
    SS & 4.09 & 4.09 & 3.79 & 3.62 & \textbf{3.61} & $-12\%$ \\
    SF & 100.00 & 100.00 & 71.43 & 57.14 & \textbf{42.86} & $-57\%$ \\
    FS & 2.21 & 2.00 & 0.88 & 0.87 & \textbf{0.71} & $-68\%$ \\
    \bottomrule
    \multicolumn{7}{l}{\footnotesize We collected 50 samples per corner (250 total for all 5 corners).}
    \end{tabular}
    }
    \vspace{-0.25in}
\end{table}
    
We measure whether training across PVT corners improves accuracy via knowledge transfer. Using the 16$\times$2 SRAM, each target corner starts with 50 of its own samples, then incrementally adds 50 samples from additional unseen corners (from target-only to all five corners).

Table~\ref{tab:global_modeling_ablation} shows that MRE decreases as the number of trained corners increases. For TT, MRE drops from 21.79\% (target-only) to 6.04\% (all corners, $-72\%$); FS improves from 2.21\% to 0.71\% ($-68\%$); SS from 4.09\% to 3.61\% ($-12\%$). The challenging SF corner benefits most, falling from 100.00\% to 42.86\% ($-57\%$). FF remains at 0.00\% across settings, indicating it is already well-modeled without additional cross-corner data.
These trends indicate that the proposed method leverages prior knowledge from correlated corners to improve accuracy and sample efficiency compared with per-corner training.

\vspace{-0.1in}
\subsection{Scalable YMCA Validation}

We evaluate YMCA on four SRAM configurations (with 144 to 1152 variational parameters). For circuits exceeding 500 dimensions (16$\times$2 and 32$\times$2), we apply sparse feature selection, compressing to approximately 48 critical dimensions. All methods use identical simulation budgets: 1000-sample budgets for small circuits, and 2500 for large; yield based on 50,000-sample MC is used as ground truth.

Table~\ref{tab:multi_corner_results_detailed} demonstrates detailed yields and predictions across all PVT corners and SRAM configs. Our method maintains high accuracy everywhere: on low-dimensional circuits (4$\times$2, 8$\times$2), it achieves best-in-class mean MRE of \textbf{0.11\%} (4$\times$2) and \textbf{0.22\%} (8$\times$2). On high-dimensional circuits where binary schemes collapse, our approach remains stable. On the 32$\times$2 (1152D) benchmark, it reaches \textbf{1.10\%} mean MRE while using fewer samples, keeping accuracy consistent across corner difficulties.
PVT combinations strongly impact yield across sizes. For 8$\times$2, yield ranges from 99.0\% (FF) down to 11.3\% (SF) and collapses to 0\% (SS); for 16$\times$2, SF is 0.14\% while FS/FF are near 100\%; for 4$\times$2, SS is 0\% despite TT/FS at 100\%; for 32$\times$2, SF is 34.9\% versus 100\% at FF/FS. Despite these swings, our method preserves accuracy across all PVTs and sizes, whereas binary baselines frequently fail at SF (e.g., 100\% error on 16$\times$2). This highlights the advantage of continuous performance modeling with learned priors in multi-corner settings.

\vspace{-0.1in}
\section{Conclusion}
\label{sec:conclusion}

We propose a novel yield analysis framework using learned priors to break the long-standing tuning barrier in yield analysis, achieving top performance with zero tuning and naturally enabling knowledge transfer to solve the YMCA challenges. Further improvements can be expected by fine-tuning TabPFN on SPICE simulation data from real circuits.

%%
%% The acknowledgments section is defined using the "acks" environment
%% (and NOT an unnumbered section). This ensures the proper
%% identification of the section in the article metadata, and the
%% consistent spelling of the heading.
% \begin{acks}
% The authors would like to thank the reviewers for their valuable feedback.
% \end{acks}

%%
%% The next two lines define the bibliography style to be used, and
%% the bibliography file.
\bibliographystyle{ACM-Reference-Format}
\bibliography{YMCA.bib}

\end{document}